\documentclass[a4paper]{article}

\usepackage{INTERSPEECH2021}
\usepackage{hyperref}
\usepackage{algorithm}      
\usepackage{amssymb}
\usepackage[noend]{algpseudocode} 

\algdef{SE}[DOWHILE]{Do}{doWhile}{\algorithmicdo}[1]{\algorithmicwhile\ #1}%

\title{SepTr: Separable Transformer for Audio Spectrogram Processing}
\name{Nicolae-C\u{a}t\u{a}lin Ristea$^{1,4}$, Radu Tudor Ionescu$^{2,*}$\thanks{$^*$corresponding author}, Fahad Shahbaz Khan$^{4, 3}$}

\address{
  $^1$University Politehnica of Bucharest, Romania\\
  $^2$University of Bucharest, Romania, $^3$Link\"{o}ping University, Sweden\\
  $^4$Mohamed bin Zayed University of Artificial Intelligence, UAE
  }
\email{r.catalin196@yahoo.ro, raducu.ionescu@gmail.com, fahad.khan@liu.se}

\begin{document}

\maketitle

\begin{abstract}
Following the successful application of vision transformers in multiple computer vision tasks, these models have drawn the attention of the signal processing community. This is because signals are often represented as spectrograms (e.g.~through Discrete Fourier Transform) which can be directly provided as input to vision transformers. However, naively applying transformers to spectrograms is suboptimal. Since the axes represent distinct dimensions, i.e.~frequency and time, we argue that a better approach is to separate the attention dedicated to each axis. To this end, we propose the \textbf{Sep}arable \textbf{Tr}ansformer (SepTr), an architecture that employs two transformer blocks in a sequential manner, the first attending to tokens within the same time interval, and the second attending to tokens within the same frequency bin. We conduct experiments on three benchmark data sets, showing that our separable architecture outperforms conventional vision transformers and other state-of-the-art methods. Unlike standard transformers, SepTr linearly scales the number of trainable parameters with the input size, thus having a lower memory footprint. Our code is available as open source at \url{https://github.com/ristea/septr}.
\end{abstract}

\setlength{\abovedisplayskip}{3pt}
\setlength{\belowdisplayskip}{3pt}

\vspace{-0.1cm}
\section{Introduction}

\begin{figure*}[!ht]
\begin{center}
\centerline{\includegraphics[width=0.82\linewidth]{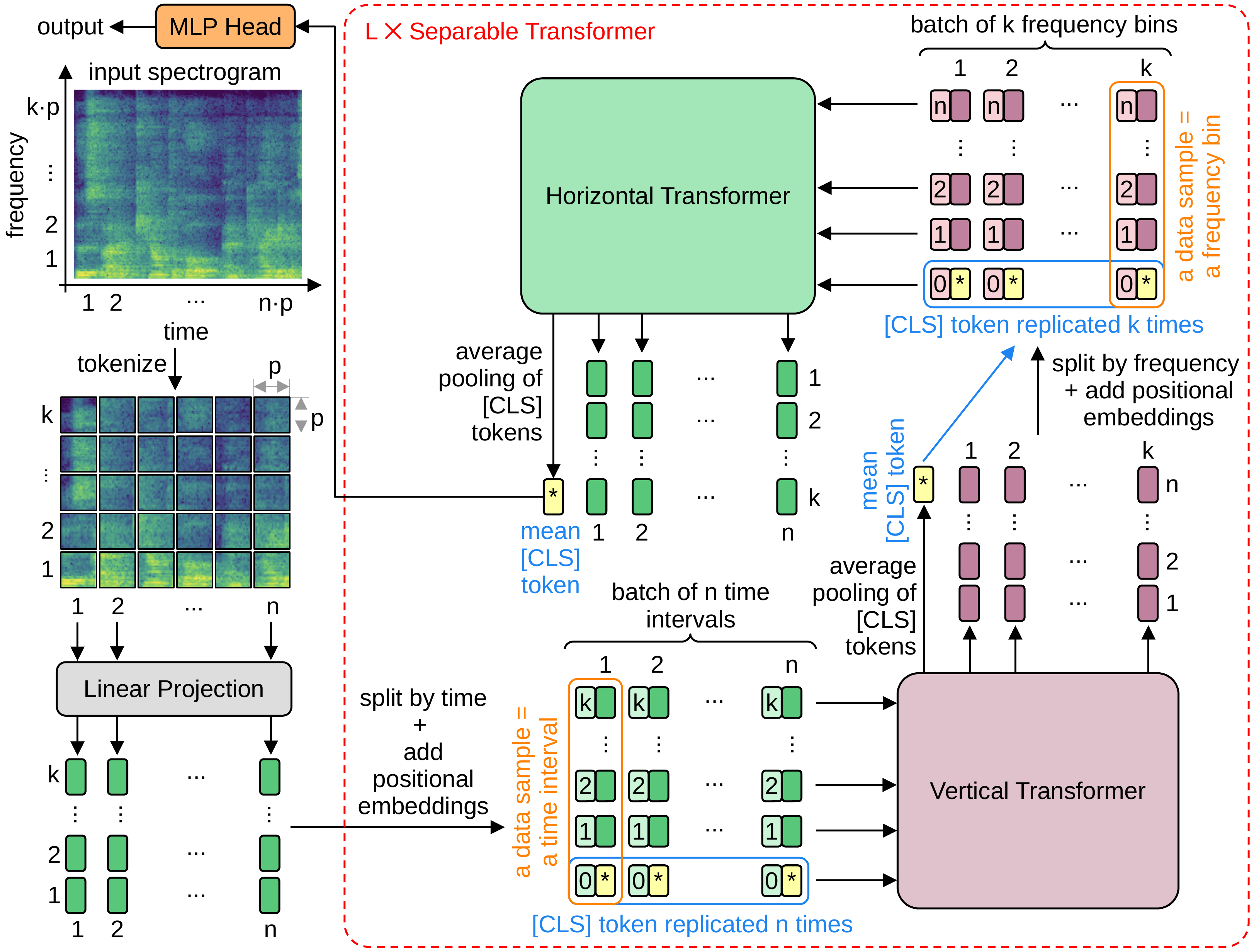}}
\vspace{-0.25cm}
\caption{Our SepTr architecture. The spectrogram is tokenized, projected and further processed by two sequential transformers (vertical and horizontal). Our separable transformer block (comprising both vertical and horizontal transformers) is repeated $L$ times, leading to a SepTr model of $L$ blocks in depth. The final class token is taken as input by an MLP head which decides the final output. 
}
\label{fig_septr}
\end{center}
\vspace{-1.0cm}
\end{figure*}

Vision transformers \cite{Carion-ECCV-2020,Chen-arXiv-2021,Dosovitskiy-ICLR-2020,Gao-MICCAI-2021,Khan-ACS-2021,Parmar-ICML-2018,Ristea-ArXiv-2021,Touvron-ICML-2021,Wu-ICCV-2021,Zheng-BMVC-2021,Zhu-ICLR-2020} have rapidly become the hottest topic in computer vision to date, showing promising results across a broad range of tasks, from object recognition \cite{Dosovitskiy-ICLR-2020,Touvron-ICML-2021,Wu-ICCV-2021} and detection \cite{Carion-ECCV-2020,Zheng-BMVC-2021,Zhu-ICLR-2020} to medical image segmentation \cite{Chen-arXiv-2021,Gao-MICCAI-2021,Hatamizadeh-WACV-2022} and generation \cite{Ristea-ArXiv-2021}. Due to the unquestionable success of vision transformers in solving many computer vision tasks, these models have also drawn the attention of the signal processing community \cite{Gong-INTERSPEECH-2021,Illium-INTERSPEECH-2021}. For example, ViT \cite{Dosovitskiy-ICLR-2020} was transferred to the processing of signals by Gong et al.~\cite{Gong-INTERSPEECH-2021}, by essentially providing visual representations of signals, i.e.~spectrograms, as input to the model. Since the axes of a spectrogram represent distinct dimensions, e.g.~frequency and time, we argue that naively applying transformers to spectrograms is suboptimal. Indeed, the attention spans across both directions, leading to a quadratic complexity with respect to the number of tokens. While operating over both horizontal and vertical axes seems reasonable for natural or medical images, we conjecture that a better approach for processing spectrograms is to separate the attention for each axis. To this end, we propose an architecture that employs two transformer blocks in a sequential manner. The first block (vertical transformer) attends to tokens within the same time interval, individually processing each time interval. Similarly, the second block (horizontal transformer) attends to tokens within the same frequency bin, independently operating over each frequency bin. To easily implement our transformer, we convert the input spectrogram into a batch of data samples along each axis, where a data sample is alternatively formed of tokens within the same time interval or within the same frequency bin, as illustrated in Figure~\ref{fig_septr}. Our approach leads to a quadratic reduction of the number of learnable parameters, which further translates into a more efficient attention mechanism. In a nutshell, our transformer separates the attention for the horizontal and vertical axes of spectrograms, thus bearing the name \textbf{Sep}arable \textbf{Tr}ansformer (SepTr).

We conduct experiments on three benchmark data sets, namely Speech Commands V2 \cite{Warden-ArXiv-2018}, ESC-50 \cite{Piczak-ACMMM-2015} and CREMA-D \cite{Cao-TAC-2014}, showing that our separable architecture attains significantly better results than models inspired by vision transformers \cite{Gong-INTERSPEECH-2021}, as well as other state-of-the-art methods \cite{georgescu2020non,He-CVPRW-2020,Kim-DCASE-2020,Majumdar-INTERSPEECH-2020,Ristea-INTERSPEECH-2021,Sailor-INTERSPEECH-2017,Shukla-ICASSP-2020}, on all benchmarks. Moreover, we demonstrate that SepTr significantly reduces the number of learnable parameters (weights) when compared to ViT.

In summary, our contribution is twofold:
\begin{itemize}
    \item \vspace{-0.1cm} We propose a novel architecture based on separable transformer blocks, which is particularly suitable for the efficient processing of spectrograms.  
    \item \vspace{-0.1cm} We provide empirical evidence to support our claims regarding the high level of effectiveness and the low number of trainable parameters of our approach in relation to competing methods.
\end{itemize}

\section{Related Work}

Due to the recent progress of attention mechanisms \cite{Vaswani-NIPS-2017}, transformers have become attractive and powerful choices for audio related tasks. On the one hand, some studies took self-attention mechanisms used in natural language processing to process text sequences and adapted them to process audio sequences and solve various audio tasks, e.g.~automatic speech recognition \cite{Kanda-INTERSPEECH-2021, Lohrenz-INTERSPEECH-2021, Leong-INTERSPEECH-2021}, speech synthesis \cite{Wu-INTERSPEECH-2021}, and speech representation \cite{Luo-INTERSPEECH-2021}. On the other hand, a few approaches adopted architectures from computer vision \cite{Gong-INTERSPEECH-2021,Illium-INTERSPEECH-2021}, mainly due to the similarity between images and time-frequency representations.

In the preliminary studies adopting transformers for audio tasks, the attention architecture was typically used in conjunction with convolutional neural networks (CNNs) \cite{Miyazaki-DCASE-2020, Kong-TASLP-2020, Gulati-INTERSPEECH-2020}. In \cite{Miyazaki-DCASE-2020, Kong-TASLP-2020}, the authors stacked a transformer on top of a CNN, while in \cite{Gulati-INTERSPEECH-2020}, the authors combined the transformer module with a CNN in each block. Our method differs from these studies in that it is convolution-free, being purely based on multi-head attention modules. 

The self-attention mechanism was widely adopted in the audio field, providing remarkable results \cite{Miyazaki-ICASSP-2020, Koizumi-ICASSP-2020}. For sound event detection, Miyazaki et al.~\cite{Miyazaki-ICASSP-2020} utilized a transformer encoder based on BERT, which consists of multiple self-attention modules, allowing the capture of both local and global contextual information of the input sequence. 
Koizumi et al.~\cite{Koizumi-ICASSP-2020} proposed a similar method based on multi-head self-attention modules, which is used in a multi-task learning setting for the speech enhancement task. They observed that the attention mechanism is able to handle long-term dependencies in speech and noise, obtaining improved results. Instead of introducing attention modules into existing architectures as \cite{Miyazaki-ICASSP-2020, Koizumi-ICASSP-2020}, we design a stand-alone transformer architecture based on separable attention, which is both effective and efficient.

Kanda et al.~\cite{Kanda-INTERSPEECH-2021} proposed an end-to-end automatic speech recognition method, which jointly performs speaker counting, speech recognition and speaker identification for monaural multi-talker audio. They replaced a memory-based approach with custom transformers, attaining better word error rates. Similarly, Wang et al.~\cite{Wang-ICASSP-2020} proposed a hybrid architecture based on transformers for speech recognition. In contrast, we design an efficient transformer based on separable attention, which exploits the data structure of time-frequency representations.

Other works \cite{Gong-INTERSPEECH-2021,Illium-INTERSPEECH-2021} relied on the similarity between digital images and time-frequency representations by applying solutions originally proposed for computer vision tasks. Illium et al.~\cite{Illium-INTERSPEECH-2021} applied the vision transformer (ViT) \cite{Dosovitskiy-ICLR-2020} to primates classification and COVID detection. Even if they naively employ the solution proposed in vision, their results surpass the baseline methods. The authors also tried to built more natural tokens for time-frequency representations, considering all the frequency bins from a specific time slot as one token. However, this approach led to inferior results compared with ViT. A more comprehensive study about the utilization of ViT in the audio domain is conducted by Gong et al.~\cite{Gong-INTERSPEECH-2021}. The authors discussed the utility of vision pre-training in the audio domain, reporting state-of-the-art results on multiple data sets. Unlike other approaches inspired by vision transformers, we propose an efficient transformer with separable attention along the axes, which is more natural for time-frequency representations, e.g.~audio spectrograms. Notably, our approach is capable of attaining superior results with a lower number of parameters. 




\section{Method}

\noindent{\bf Data representation.}
We transform each audio sample into a 2D time-frequency matrix, obtaining an image-like representation. To this end, we compute the discrete Short Time Fourier Transform (STFT), as follows:
\begin{equation}\label{eq_stft}
\mbox{STFT}\{x[n]\}(m, k)=\!\!\sum_{n=-\infty}^{\infty}\!\! x[n] \cdot w[n-m R] \cdot e^{-j \frac{2 \pi}{N_x}k n},
\end{equation}
where $x[n]$ is the discrete input signal, $w[n]$ is the window function (in our approach, Hamming), $N_x$ is the STFT length and $R$ is the hop size. 
Next, we compute the spectrogram as the squared magnitude of the STFT and map the frequency bins onto the Mel scale.

\noindent{\bf Overview of our architecture.}
We propose SepTr, a \textbf{Sep}arable \textbf{Tr}ansformer architecture composed of two sequential transformer blocks, each attending to tokens within separate dimensions. The architecture does not impose a certain axis (time or frequency) for the first transformer block, being flexible in this regard. Without loss of generality, in Figure~\ref{fig_septr}, we illustrate a model that separates the tokens along the time axis first. Our separable transformer block can be repeated $L$ times to increase the depth of the architecture. The final mean class token (given by the last separable block) is processed by a multi-layer perceptron (MLP), which outputs the final prediction of our model. We next present the building blocks of SepTr.

\noindent{\bf Tokenization and linear projection.}
Considering a spectrogram with $k\cdot p \in \mathbb{N}$ frequency bins and $n\cdot p \in \mathbb{N}$ time slots as input, in the first stage, we divide the input spectrogram into $k\cdot n$ square patches (tokens) of size $p \times p$. 
In the experiments, we take patches of size $1 \times 1$ as individual tokens, since we consider that it is more natural to compute the self-attention at the finest possible level.
However, for a better visualization of the architecture shown in Figure \ref{fig_septr}, the tokens are depicted as larger patches.
Regardless of their size, the tokens are further fed into a linear projection block, which projects the tokens into $d$-dimensional vectors. Let $T \in \mathbb{R}^{n \times k \times d}$ represent the output tensor of the linear projection layer, where $\textbf{T}_{i, j} \in \mathbb{R}^d$ is a projected token, $\forall i \in \{1,2,...,k\}$ and $j \in \{1,2,...,n\}$, where $d \in \mathbb{N}$ is the latent token dimension. 

\noindent{\bf Vertical transformer.}
We separate the projected tokens in the time domain into data sub-samples denoted as $\textbf{T}_{:,j}=\left[\textbf{T}_{1, j},\textbf{T}_{2, j},...,\textbf{T}_{k, j}  \right] \in \mathbb{R}^{k \times d}$, thus obtaining a batch of $n$ data samples (one sample for each time slot $j$), where each data sample is composed of $k$ tokens. Moreover, we replicate the class token $\textbf{T}_{\scriptsize{\mbox{[CLS]}}} \in \mathbb{R}^d$ $n$ times and add one copy to each data sample $\textbf{T}_{:,j}$. We also add a learnable positional embedding to each token. The resulting batch of $n$ individual data samples is further processed by the vertical transformer. 

\noindent{\bf Horizontal transformer.}
Let $\textbf{T}^V_{:,j}$ denote a data sample given as output by the vertical transformer. Next, we concatenate the processed data samples into a tensor $\textbf{T}^V \in \mathbb{R}^{n \times k \times d}$, while decoupling the class tokens and applying average pooling to obtain a mean class token denoted as $\hat{\textbf{T}}^V_{\scriptsize{\mbox{[CLS]}}}$. We proceed by separating the tensor $\textbf{T}^V$ in the frequency domain into data sub-samples denoted as $\textbf{T}^V_{i,:}=\left[\textbf{T}^V_{i, 1},\textbf{T}^V_{i, 2},...,\textbf{T}^V_{i, n}  \right] \in \mathbb{R}^{n \times d}$, generating a batch of $k$ data samples (one sample for each frequency bin $i$), where each sample is formed of $n$ tokens. As for the vertical transformer, we replicate the class token $\hat{\textbf{T}}^V_{\scriptsize{\mbox{[CLS]}}}$ $k$ times and add one copy to each data sample $\textbf{T}^V_{i,:}$. We further append a learnable positional embedding to each token. The resulting batch of $k$ data samples is given as input to the horizontal transformer.


\noindent{\bf Transformer block.}
The operations performed inside the vertical and horizontal transformers are identical, the only difference being the format of the input data samples. We thus describe the inner operations for the general case. Let $\textbf{X} \in \mathbb{R}^{m \times d}$ denote a sequence of $m$ tokens (either $\textbf{T}_{:,j}$ or $\textbf{T}^V_{i,:}$), where $m \in \{k,n\}$ and $d$ is the embedding dimension of each token. Let $f$ be a multi-head attention layer, $g$ a multi-layer perceptron, $\mbox{\emph{norm}}$ a normalization layer, and $\textbf{P}, \textbf{R} \in \mathbb{R}^{m \times d}$ some auxiliary tensors. The transformer block is formally described as follows:
\begin{equation}
\textbf{P} = f(\mbox{\emph{norm}}(\textbf{X})) + \textbf{X},
\end{equation}
\begin{equation}
\textbf{R} = g(\mbox{\emph{norm}}(\textbf{P}) ) + \textbf{P}.
\end{equation}

The goal of the transformer block is to capture the interaction among all $m$ entities by encoding each entity in terms of the global contextual information. This is achieved via the multi-head attention layer $f$. This layer comprises three learnable weight matrices ($\textbf{W}^Q \in \mathbb{R}^{d \times d_q}$, $\textbf{W}^K \in \mathbb{R}^{d \times d_k}$, $\textbf{W}^V \in \mathbb{R}^{d \times d_v}$, where $d_q = d_k$) which are used to derive the queries $\textbf{Q}$, keys $\textbf{K}$ and values $\textbf{V}$ from the input sequence $\textbf{X}$. Indeed, the input sequence $\textbf{X}$ is
first projected onto these weight matrices to get $\textbf{Q} = \textbf{X}\cdot\textbf{W}^Q$, $\textbf{K} = \textbf{X}\cdot\textbf{W}^K$, and $\textbf{V} = \textbf{X}\cdot\textbf{W}^V$, respectively. The output $\textbf{Z} \in \mathbb{R}^{d \times d_v}$ of the self-attention is given by:
\begin{equation}
    \textbf{Z} = \mbox{\emph{softmax}}\left( \frac{\textbf{Q}\cdot\textbf{K}'}{\sqrt{d_q}}\right)\cdot \textbf{V},
\end{equation}
where $\textbf{K}'$ is the transpose of $\textbf{K}$. 


\section{Experiments}

\subsection{Data sets}

\noindent{\bf ESC-50.}
The ESC-50 \cite{Piczak-ACMMM-2015} data set is a collection of 2,000 samples of 5 seconds each, comprising 50 classes of various common sound events. Samples are recorded at a 44.1 kHz sampling frequency, with a single channel. For evaluation, we followed the 5-fold cross-validation procedure described in \cite{Gong-INTERSPEECH-2021}.

\noindent{\bf Speech Commands V2.}
The Speech Commands V2 (SCV2) \cite{Warden-ArXiv-2018} data set is composed of spoken words, being designed to train and evaluate keyword spotting systems. It consists of 105,829 1-second recordings of 35 common speech commands. We used the official training (84,843 samples), validation (9,981 samples) and test (11,005 samples) splits proposed in \cite{Warden-ArXiv-2018}.

\noindent{\bf CREMA-D.}
The CREMA-D multi-modal database \cite{Cao-TAC-2014} is formed of 7,442 videos of 91 actors (48 male and 43 female) of different ethnic groups. The actors perform various emotions while uttering 12 particular sentences that evoke one of the 6 emotion categories. 
Following \cite{Ristea-INTERSPEECH-2021}, we conduct experiments only on the audio modality, dividing the audio samples into $70\%$ for training, $15\%$ for validation and $15\%$ for testing.

\subsection{Evaluation setup}

\noindent{\bf Performance metrics.}
In all our experiments, we employ the classification accuracy as evaluation measure. We also conduct significance tests to compare SepTr with the top competitor, using a paired McNemar's test \cite{Dietterich-NC-1998} at a significance level of 0.01.

\noindent{\bf Data preprocessing.} 
For CREMA-D, we first standardize all audio clips to a fixed dimension of 4 seconds by padding or clipping the samples. For both CREMA-D and SCV2, we apply the STFT with $N_x\!=\!1024$, $R\!=\!64$ and a window size of 512. We use the same $N_x$ value and window length for ESC-50, but we increase the hop size to $R\!=\!128$. Next, for each STFT, we compute the square root of the magnitude and map the values to 128 Mel bins. The result is converted to a logarithmic scale (decibels) and normalized to the interval $[0,1]$, generating a single-channel output matrix. In all our experiments, we use the following data augmentation methods: noise perturbation, time shifting, speed perturbation, mix-up and SpecAugment \cite{Park-INTERSPEECH-2019}.

\noindent{\bf Hyperparameter tuning.} 
While using the validation sets to tune the hyperparameters of SepTr and ViT, we found common optimal hyperparameters for all data sets. Thus, all models are optimized with Adam using the cross-entropy loss function. We start with an initial learning rate of $10^{-4}$ and use a decay factor of $0.5$ after every 10 epochs. We train each model for 50 epochs on mini-batches of 4 samples. For SepTr, we set the number of blocks to $L\!=\!3$ and the token size to $d\!=\!256$, while for ViT we use a depth of $L\!=\!6$ and the same token size ($d\!=\!256$). Since our transformer block comprises two (vertical and horizontal) attention modules, the depths of SepTr and ViT are equivalent. Both SepTr and ViT are designed with $5$ attention heads. For the ViT architecture, we divide the spectrograms into $8 \times 8$ patches at an overlap of $6$ pixels in both directions, following \cite{Gong-INTERSPEECH-2021}. We hereby underline that we were not able to train the ViT architecture with $1\times 1$ patches due to out-of-memory issues, while SepTR does not suffer from this problem (see Figure~\ref{fig_noparams}).

\begin{table}[!t]
\caption{Results of SepTr and its ablated versions in comparison with various state-of-the-art methods on CREMA-D. Significantly better results compared to \cite{Ristea-INTERSPEECH-2021} are marked with $\dagger$, using a paired McNemar's test \cite{Dietterich-NC-1998} at a significance level of $0.01$. Top scores are highlighted in bold.}
\label{tab_results_cremad}
\vspace{-0.2cm}
\centering
\begin{tabular}{l c}
\toprule
{\textbf{Method}} & {\textbf{Accuracy}}\\
\midrule
GRU (Shukla et al.~\cite{Shukla-ICASSP-2020}) & 55.01\% \\
GAN (He et al.~\cite{He-CVPRW-2020}) & 58.71\% \\
ResNet-18 (Georgescu et al.~\cite{georgescu2020non}) & 65.15\% \\
ResNet-18 ensemble (Ristea et al.~\cite{Ristea-INTERSPEECH-2021}) & 68.12\% \\
ViT (Gong et al.~\cite{Gong-INTERSPEECH-2021})  & 67.81\% \\
\midrule
SepTr-V         & 65.29\% \\
SepTr-H         & 65.11\% \\
\midrule
SepTr-HV         & 70.31\%$^\dagger$ \\
SepTr-VH (proposed)        & \textbf{70.47\%}$^\dagger$ \\
\bottomrule
\end{tabular}
\vspace{-0.1cm}
\end{table}

\subsection{Ablation study}

We first analyze the impact of the vertical and horizontal transformers, when these modules are used alone or jointly. We present ablation results on the CREMA-D data set in Table \ref{tab_results_cremad}. We include ablated versions of SepTr performing attention either inside individual time slots (SepTr-V uses only vertical transformer blocks) or frequency bins (SepTr-H uses only horizontal transformer blocks). We observe that computing the attention on a single axis with SepTr-V or SepTr-H leads to suboptimal results, even below the ViT baseline. This shows that performing attention on one axis only is not sufficient.

Further, we show results while alternating the order of the vertical and horizontal transformers, i.e.~SepTr-HV starts with a horizontal transformer and SepTr-VH starts with a vertical transformer. Regardless of how the separable attention is applied via SepTr-HV or SepTr-VH, we observe important performance boosts (higher than $5\%$) over the models attending to one of the axes (SepTr-V and SepTr-H). Switching the order of the vertical and horizontal transformers has a marginal influence on the accuracy level, both SepTr-HV or SepTr-VH being able to outperform the state-of-the-art methods by significant margins. As we obtained slightly better results with SepTr-VH, we continued the rest of the experiments with this architecture, which we further simply refer to as \emph{SepTr}.

\begin{table}[!t]
\caption{Results of SepTr versus various state-of-the-art methods on Speech Commands V2 (SCV2) and ESC-50. Significantly better results compared to ViT \cite{Gong-INTERSPEECH-2021} are marked with $\dagger$, using a paired McNemar's test \cite{Dietterich-NC-1998} at a significance level of $0.01$. Top scores are highlighted in bold.}
\label{tab_results_2}
\setlength\tabcolsep{4.0pt}  
\vspace{-0.2cm}
\centering
\begin{tabular}{l c c}
\toprule
{\textbf{Method}} 
& \textbf{SCV2} & {\textbf{ESC-50}}\\
\midrule
RBM (Sailor et al.~\cite{Sailor-INTERSPEECH-2017})  & - &  86.50\% \\
EfficientNet (Kim et al.~\cite{Kim-DCASE-2020}) & - & 89.50\% \\
MatchboxNet (Majumdar et al.~\cite{Majumdar-INTERSPEECH-2020}) & 97.40\% & - \\
ViT (Gong et al.~\cite{Gong-INTERSPEECH-2021}) & 98.11\% & 88.70\% \\
\midrule
SepTr (proposed)         & \textbf{98.51\%}$^\dagger$ & \textbf{91.13\%}$^\dagger$\\
\bottomrule
\end{tabular}
\vspace{-0.1cm}
\end{table}

\subsection{Results}

\noindent{\bf Effectiveness.}
On the CREMA-D data set (see Table~\ref{tab_results_cremad}), we obtain state-of-the-art results, reaching an accuracy of $70.47\%$ and surpassing the previous best method \cite{Ristea-INTERSPEECH-2021}, composed of an ensemble of ResNet models, by $2.35\%$. Moreover, we attain a performance boost of $2.66\%$ compared to ViT \cite{Gong-INTERSPEECH-2021}.
On the SCV2 and ESC-50 data sets (see Table~\ref{tab_results_2}), we consider state-of-the-art methods that are not pre-trained on external data, ensuring a fair comparison with SepTr, which is trained from scratch. We observe that SepTr attains superior results on both data sets. On SCV2, SepTr attains an accuracy of $98.51\%$, which is $0.40\%$ higher than the previous state-of-the-art accuracy obtained by ViT \cite{Gong-INTERSPEECH-2021}. On ESC-50, SepTr yields an accuracy of $91.13\%$, surpassing ViT \cite{Gong-INTERSPEECH-2021} by $2.43\%$ and EfficientNet \cite{Kim-DCASE-2020} by $1.63\%$, respectively. In summary, SepTr surpasses all the state-of-the-art methods on each and every data set by significant margins.

\begin{figure}
\begin{center}
\centerline{\includegraphics[width=0.94\linewidth]{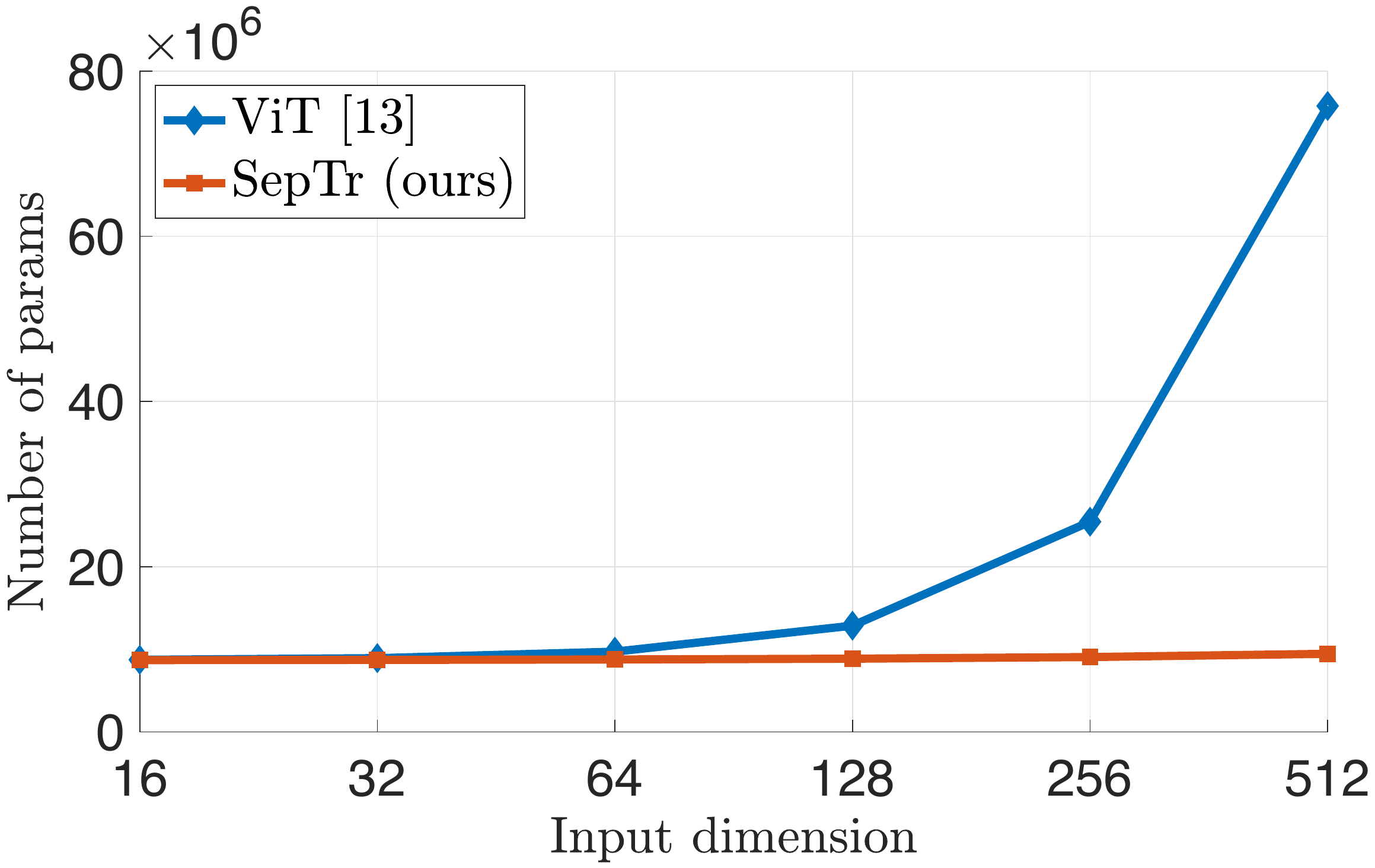}}
\vspace{-0.2cm}
\caption{The number of trainable parameters for SepTr and ViT \cite{Gong-INTERSPEECH-2021} with respect to the input dimension.}
\label{fig_noparams}
\vspace{-1.0cm}
\end{center}
\end{figure}

\noindent{\bf Efficiency.}
We compare the performance of SepTr and ViT in terms of the memory footprint (number of learnable parameters), inference time and number of multiply-accumulate operations (MACs), on an Nvidia GeForce GTX 3090 GPU with 24 GB of VRAM. We compare equivalent models having 6 attention layers each, which are fed with $1 \times 1$ patches. In terms of GMACs ($80.57$ for SepTr versus $80.09$ for ViT) and inference time ($5.72$ ms for SepTr versus $5.69$ ms for ViT), the differences are negligible, but when we refer to the memory footprint, we observe a large difference between the two models. In Figure \ref{fig_noparams}, we illustrate the number of learnable parameters as a function of the input spectrogram size. 
We observe that SepTr has an almost constant number of parameters with respect to the input dimension, while the number of weights in ViT exhibits a quadratic growth, leading to comparatively larger models. For instance, for an input size of $512 \times 512$, SepTr has 9.4M parameters, while ViT has 75.7M. Unlike ViT, our model is able to handle high resolution spectrograms, an essential advantage leading to high accuracy levels in the audio domain.

\section{Conclusion}
In this work, we proposed a novel Separable Transformer (SepTr) architecture composed of two sequential transformers, each computing the attention on a different axis of the input spectrogram, either time or frequency. We obtained state-of-the-art results on three data sets, surpassing all competing methods by statistically significant margins. Moreover, we showed that SepTr reduces the memory footprint when compared with ViT, without impacting the inference time or MACs. In future work, we aim to employ SepTr for other signal processing tasks.

\vspace{0.2cm}
\noindent
{\bf Acknowledgements.} Work supported by a grant of the Romanian Ministry of Education and Research, CNCS - UEFISCDI, project no. PN-III-P1-1.1-TE-2019-0235, within PNCDI III.

\bibliographystyle{IEEEtran}
\bibliography{mybib}

\begin{thebibliography}{10}
\providecommand{\url}[1]{#1}
\csname url@samestyle\endcsname
\providecommand{\newblock}{\relax}
\providecommand{\bibinfo}[2]{#2}
\providecommand{\BIBentrySTDinterwordspacing}{\spaceskip=0pt\relax}
\providecommand{\BIBentryALTinterwordstretchfactor}{4}
\providecommand{\BIBentryALTinterwordspacing}{\spaceskip=\fontdimen2\font plus
\BIBentryALTinterwordstretchfactor\fontdimen3\font minus
  \fontdimen4\font\relax}
\providecommand{\BIBforeignlanguage}[2]{{%
\expandafter\ifx\csname l@#1\endcsname\relax
\typeout{** WARNING: IEEEtran.bst: No hyphenation pattern has been}%
\typeout{** loaded for the language `#1'. Using the pattern for}%
\typeout{** the default language instead.}%
\else
\language=\csname l@#1\endcsname
\fi
#2}}
\providecommand{\BIBdecl}{\relax}
\BIBdecl

\bibitem{Carion-ECCV-2020}
N.~Carion, F.~Massa, G.~Synnaeve, N.~Usunier, A.~Kirillov, and S.~Zagoruyko,
  ``End-to-end object detection with transformers,'' in \emph{Proceedings of
  ECCV}.\hskip 1em plus 0.5em minus 0.4em\relax Springer, 2020, pp. 213--229.

\bibitem{Chen-arXiv-2021}
J.~Chen, Y.~Lu, Q.~Yu, X.~Luo, E.~Adeli, Y.~Wang, L.~Lu, A.~L. Yuille, and
  Y.~Zhou, ``{TransUNet: Transformers Make Strong Encoders for Medical Image
  Segmentation},'' \emph{arXiv preprint arXiv:2102.04306}, 2021.

\bibitem{Dosovitskiy-ICLR-2020}
A.~Dosovitskiy, L.~Beyer, A.~Kolesnikov, D.~Weissenborn, X.~Zhai,
  T.~Unterthiner, M.~Dehghani, M.~Minderer, G.~Heigold, S.~Gelly \emph{et~al.},
  ``An image is worth 16x16 words: Transformers for image recognition at
  scale,'' in \emph{Proceedings of ICLR}, 2021.

\bibitem{Gao-MICCAI-2021}
Y.~Gao, M.~Zhou, and D.~Metaxas, ``{UTNet: A Hybrid Transformer Architecture
  for Medical Image Segmentation},'' in \emph{Proceedings of MICCAI}, 2021, pp.
  61--71.

\bibitem{Khan-ACS-2021}
S.~Khan, M.~Naseer, M.~Hayat, S.~W. Zamir, F.~S. Khan, and M.~Shah,
  ``{Transformers in Vision: A Survey},'' \emph{ACM Computing Surveys}, 2021.

\bibitem{Parmar-ICML-2018}
N.~Parmar, A.~Vaswani, J.~Uszkoreit, L.~Kaiser, N.~Shazeer, A.~Ku, and D.~Tran,
  ``Image transformer,'' in \emph{Proceedings of ICML}, 2018, pp. 4055--4064.

\bibitem{Ristea-ArXiv-2021}
N.-C. Ristea, A.-I. Miron, O.~Savencu, M.-I. Georgescu, N.~Verga, F.~S. Khan,
  and R.~T. Ionescu, ``{CyTran: Cycle-Consistent Transformers for Non-Contrast
  to Contrast CT Translation},'' \emph{arXiv preprint arXiv:2110.06400}, 2021.

\bibitem{Touvron-ICML-2021}
H.~Touvron, M.~Cord, M.~Douze, F.~Massa, A.~Sablayrolles, and H.~J{\'e}gou,
  ``Training data-efficient image transformers \& distillation through
  attention,'' in \emph{Proceedings of ICML}, 2021, pp. 10\,347--10\,357.

\bibitem{Wu-ICCV-2021}
H.~Wu, B.~Xiao, N.~Codella, M.~Liu, X.~Dai, L.~Yuan, and L.~Zhang, ``{CvT:
  Introducing Convolutions to Vision Transformers},'' in \emph{Proceedings of
  ICCV}, 2021, pp. 22--31.

\bibitem{Zheng-BMVC-2021}
M.~Zheng, P.~Gao, X.~Wang, H.~Li, and H.~Dong, ``End-to-end object detection
  with adaptive clustering transformer,'' in \emph{Proceedings of BMVC}, 2020.

\bibitem{Zhu-ICLR-2020}
X.~Zhu, W.~Su, L.~Lu, B.~Li, X.~Wang, and J.~Dai, ``Deformable detr: Deformable
  transformers for end-to-end object detection,'' in \emph{Proceedings of
  ICLR}, 2020.

\bibitem{Hatamizadeh-WACV-2022}
A.~Hatamizadeh, D.~Yang, H.~Roth, and D.~Xu, ``{UNETR: Transformers for 3D
  Medical Image Segmentation},'' in \emph{Proceedings of WACV}, 2022, pp.
  574--584.

\bibitem{Gong-INTERSPEECH-2021}
Y.~Gong, Y.-A. Chung, and J.~Glass, ``{AST: Audio Spectrogram Transformer},''
  in \emph{Proceedings of INTERSPEECH}, 2021, pp. 571--575.

\bibitem{Illium-INTERSPEECH-2021}
S.~Illium, R.~M\"u{}ller, A.~Sedlmeier, and C.-L. Popien, ``{Visual
  Transformers for Primates Classification and Covid Detection},'' in
  \emph{Proceedings of INTERSPEECH}, 2021, pp. 451--455.

\bibitem{Warden-ArXiv-2018}
P.~Warden, ``{Speech Commands: A Dataset for Limited-Vocabulary Speech
  Recognition},'' \emph{arXiv preprint arXiv:1804.03209}, 2018.

\bibitem{Piczak-ACMMM-2015}
K.~J. Piczak, ``{ESC: Dataset for Environmental Sound Classification},'' in
  \emph{Proceedings of ACMMM}, 2015, pp. 1015--1018.

\bibitem{Cao-TAC-2014}
H.~Cao, D.~G. Cooper, M.~K. Keutmann, R.~C. Gur, A.~Nenkova, and R.~Verma,
  ``{CREMA-D: Crowd-sourced emotional multimodal actors dataset},'' \emph{IEEE
  Transactions on Affective Computing}, vol.~5, no.~4, pp. 377--390, 2014.

\bibitem{georgescu2020non}
M.-I. Georgescu, R.~T. Ionescu, N.-C. Ristea, and N.~Sebe, ``{Non-linear
  Neurons with Human-like Apical Dendrite Activations},'' \emph{arXiv preprint
  arXiv:2003.03229}, 2020.

\bibitem{He-CVPRW-2020}
G.~He, X.~Liu, F.~Fan, and J.~You, ``{Image2Audio: Facilitating Semi-supervised
  Audio Emotion Recognition with Facial Expression Image},'' in
  \emph{Proceedings of CVPRW}, 2020, pp. 912--913.

\bibitem{Kim-DCASE-2020}
J.~Kim, ``Urban sound tagging using multi-channel audio feature with
  convolutional neural networks,'' \emph{Proceedings of DCASE}, 2020.

\bibitem{Majumdar-INTERSPEECH-2020}
S.~Majumdar and B.~Ginsburg, ``{MatchboxNet: 1D Time-Channel Separable
  Convolutional Neural Network Architecture for Speech Commands Recognition},''
  in \emph{Proceedings of INTERSPEECH}, 2020, pp. 3356--3360.

\bibitem{Ristea-INTERSPEECH-2021}
N.-C. Ristea and R.~T. Ionescu, ``Self-paced ensemble learning for speech and
  audio classification,'' in \emph{Proceedings of INTERSPEECH}, 2021, pp.
  2836--2840.

\bibitem{Sailor-INTERSPEECH-2017}
H.~B. Sailor, D.~M. Agrawal, and H.~A. Patil, ``{Unsupervised Filterbank
  Learning Using Convolutional Restricted Boltzmann Machine for Environmental
  Sound Classification},'' in \emph{Proceedings of INTERSPEECH}, vol.~8, 2017,
  p.~9.

\bibitem{Shukla-ICASSP-2020}
A.~Shukla, K.~Vougioukas, P.~Ma, S.~Petridis, and M.~Pantic, ``Visually guided
  self-supervised learning of speech representations,'' in \emph{Proceedings of
  ICASSP}, 2020, pp. 6299--6303.

\bibitem{Vaswani-NIPS-2017}
A.~Vaswani, N.~Shazeer, N.~Parmar, J.~Uszkoreit, L.~Jones, A.~N. Gomez,
  {\L}.~Kaiser, and I.~Polosukhin, ``Attention is all you need,'' in
  \emph{Proceedings of NIPS}, 2017, pp. 5998--6008.

\bibitem{Kanda-INTERSPEECH-2021}
N.~Kanda, G.~Ye, Y.~Gaur, X.~Wang, Z.~Meng, Z.~Chen, and T.~Yoshioka,
  ``{End-to-End Speaker-Attributed ASR with Transformer},'' in
  \emph{Proceedings of INTERSPEECH}, 2021, pp. 4413--4417.

\bibitem{Lohrenz-INTERSPEECH-2021}
T.~Lohrenz, Z.~Li, and T.~Fingscheidt, ``{Multi-Encoder Learning and Stream
  Fusion for Transformer-Based End-to-End Automatic Speech Recognition},'' in
  \emph{Proceedings of INTERSPEECH}, 2021, pp. 2846--2850.

\bibitem{Leong-INTERSPEECH-2021}
C.-H. Leong, Y.-H. Huang, and J.-T. Chien, ``{Online Compressive Transformer
  for End-to-End Speech Recognition},'' in \emph{Proceedings of INTERSPEECH},
  2021, pp. 2082--2086.

\bibitem{Wu-INTERSPEECH-2021}
C.~Wu, Z.~Xiu, Y.~Shi, O.~Kalinli, C.~Fuegen, T.~Koehler, and Q.~He,
  ``{Transformer-based Acoustic Modeling for Streaming Speech Synthesis},'' in
  \emph{Proceedings of INTERSPEECH}, 2021, pp. 146--150.

\bibitem{Luo-INTERSPEECH-2021}
J.~Luo, J.~Wang, N.~Cheng, and J.~Xiao, ``{Dropout Regularization for
  Self-Supervised Learning of Transformer Encoder Speech Representation},'' in
  \emph{Proceedings of INTERSPEECH}, 2021, pp. 1169--1173.

\bibitem{Miyazaki-DCASE-2020}
K.~Miyazaki, T.~Komatsu, T.~Hayashi, S.~Watanabe, T.~Toda, and K.~Takeda,
  ``Convolution augmented transformer for semi-supervised sound event
  detection,'' in \emph{Proceedings of DCASE}, 2020, pp. 100--104.

\bibitem{Kong-TASLP-2020}
Q.~Kong, Y.~Xu, W.~Wang, and M.~D. Plumbley, ``{Sound event detection of weakly
  labelled data with CNN-transformer and automatic threshold optimization},''
  \emph{IEEE/ACM Transactions on Audio, Speech, and Language Processing},
  vol.~28, pp. 2450--2460, 2020.

\bibitem{Gulati-INTERSPEECH-2020}
A.~Gulati, J.~Qin, C.-C. Chiu, N.~Parmar, Y.~Zhang, J.~Yu, W.~Han, S.~Wang,
  Z.~Zhang, Y.~Wu \emph{et~al.}, ``Conformer: Convolution-augmented transformer
  for speech recognition,'' in \emph{Proceedings of INTERSPEECH}, 2020, pp.
  5036--5040.

\bibitem{Miyazaki-ICASSP-2020}
K.~Miyazaki, T.~Komatsu, T.~Hayashi, S.~Watanabe, T.~Toda, and K.~Takeda,
  ``Weakly-supervised sound event detection with self-attention,'' in
  \emph{Proceedings of ICASSP}, 2020, pp. 66--70.

\bibitem{Koizumi-ICASSP-2020}
Y.~Koizumi, K.~Yatabe, M.~Delcroix, Y.~Masuyama, and D.~Takeuchi, ``Speech
  enhancement using self-adaptation and multi-head self-attention,'' in
  \emph{Proceedings of ICASSP}, 2020, pp. 181--185.

\bibitem{Wang-ICASSP-2020}
Y.~Wang, A.~Mohamed, D.~Le, C.~Liu, A.~Xiao, J.~Mahadeokar, H.~Huang,
  A.~Tjandra, X.~Zhang, F.~Zhang \emph{et~al.}, ``Transformer-based acoustic
  modeling for hybrid speech recognition,'' in \emph{Proceedings of ICASSP},
  2020, pp. 6874--6878.

\bibitem{Dietterich-NC-1998}
T.~G. Dietterich, ``{Approximate Statistical Tests for Comparing Supervised
  Classification Learning Algorithms},'' \emph{Neural Computation}, vol.~10,
  no.~7, pp. 1895--1923, 1998.

\bibitem{Park-INTERSPEECH-2019}
D.~S. Park, W.~Chan, Y.~Zhang, C.-C. Chiu, B.~Zoph, E.~D. Cubuk, and Q.~V. Le,
  ``{SpecAugment: A Simple Data Augmentation Method for Automatic Speech
  Recognition},'' in \emph{Proceedings of INTERSPEECH}, 2019, pp. 2613--2617.

\end{thebibliography}

\end{document}